\documentclass{article}

 \usepackage[dblblindworkshop, final]{neurips_2025}
\workshoptitle{Mechanistic Interpretability Workshop}

\usepackage[utf8]{inputenc} 
\usepackage[T1]{fontenc}    
\usepackage{hyperref}       
\usepackage{url}            
\usepackage{booktabs}       
\usepackage{amsfonts}       
\usepackage{nicefrac}       
\usepackage{microtype}      
\usepackage{xcolor}         

\title{LLM Probing with Contrastive Eigenproblems: Improving Understanding and Applicability of CCS}

\author{%
  Stefan~F.~Schouten \\
  Vrije Universiteit Amsterdam\\
  \texttt{s.f.schouten@vu.nl} \\
  \And
  Peter~Bloem \\
  Vrije Universiteit Amsterdam\\
  \texttt{p.bloem@vu.nl} \\
}

\usepackage{placeins}

\usepackage{mathtools}
\usepackage{amsmath}
\usepackage{amsthm}
\usepackage{bm}
\usepackage{amssymb}
\usepackage{graphicx}
\usepackage{relsize}        
\usepackage{stackengine}
\stackMath

\usepackage{booktabs} 
\usepackage{multirow}

\usepackage{subcaption}
\usepackage{wrapfig}

\usepackage{pgfplots}
\usepackage{pgfmath}
\usepackage{tikz}
\usetikzlibrary{calc}

\usepackage{todonotes}

\usepackage{cleveref}

\usepackage{listingsutf8}

\usepackage[normalem]{ulem}

\let\vec\mathbf
\let\tp\intercal

\DeclareMathOperator*{\argmin}{arg\,min}
\DeclareMathOperator*{\argmax}{arg\,max}
\newcommand{\E}{\mathbb{E}}

\newcommand{\xp}{\vec{x}^+}
\newcommand{\xm}{\vec{x}^-}
\newcommand{\Xp}[1][]{\stackon[0.3ex]{\vec{X}#1}{\scriptscriptstyle \smash{\hspace{.06ex}\bm{+}}\vphantom{=}}}

\newcommand{\Xm}[1][]{\stackon[0.28ex]{\vec{X}#1}{\scriptscriptstyle \smash{\bm{-}}\vphantom{=}}}

\newcommand{\Xpm}[1][]{\stackon[0.3ex]{\vec{X}#1}{\scriptscriptstyle \mathclap{\smash{\bm{+\kern-0.15ex-}}\vphantom{=}}}}

\let\autoref\Cref

\def\dir{\bm\theta}
\def\udir{\hat{\bm\theta}}

\newcommand{\MatchBracketsM}[1]{%
  \ifx\relax#1\empty
  \else
    #1 &
    \\\relax
    \expandafter\MatchBracketsM
  \fi
}
\newcommand{\ManyVPhantom}[1]{
  \ifx\relax#1\empty
  \else
    \vphantom{#1}
    \expandafter\ManyVPhantom
  \fi
}

\begin{document}

\maketitle

\begin{abstract}
Contrast-Consistent Search (CCS) is an unsupervised probing method able to test whether large language models represent binary features, such as sentence truth, in their internal activations. 
While CCS has shown promise, its two-term objective has been only partially understood. 
In this work, we revisit CCS with the aim of clarifying its mechanisms and extending its applicability. 
We argue that what should be optimized for, is \textit{relative} contrast consistency. 
Building on this insight, we reformulate CCS as an eigenproblem, yielding closed-form solutions with interpretable eigenvalues and natural extensions to multiple variables. 
We evaluate these approaches across a range of datasets, finding that they recover similar performance to CCS, while avoiding problems around sensitivity to random initialization.
Our results suggest that relativizing contrast consistency not only improves our understanding of CCS but also opens pathways for broader probing and mechanistic interpretability methods.




\end{abstract}

\section{Introduction}
When Large Language Models (LLMs) perform well on benchmarks for a given domain or task, the results are sometimes questioned; in part because of a limited understanding of their working. 
How is it that LLMs do what they do? 
Without a clear picture of how an LLM approaches its tasks, we cannot verify if that approach is sensible, or how well it will do outside of benchmarks.
The goal of Mechanistic Interpretability is to remedy this situation by identifying both:
(1) what mechanisms are responsible for model behaviors; and, 
(2) what variables those mechanisms use, where they are encoded, and if they correspond to interpretable \textit{features}.

This paper provides an in-depth look at Contrast-Consistent Search (CCS) \citep{burns_discovering_2023}.
This unsupervised probing method was introduced to determine if language models represent sentences as true or false.
Being unsupervised, it has one advantage: it does not assume that the model's truth-values agree with human-authored labels.
CCS has a two-termed loss function designed to find a parameter vector that makes the probe assign probabilities to sentences and their negations which add up to one.
We perform an ablation of the method's loss terms and find that one of the terms is necessary for a different reason than what originally motivated its inclusion.
We argue that contrast consistency should be defined in a \textit{relative} way.
Based on this insight, we find that CCS's objective can be made completely linear. 
This allows us to solve for contrast consistent directions using \textit{Contrastive Eigenproblems}. 
This approach yields interpretable eigenvalues that provide additional insights. 
We demonstrate this by showing that datasets where CCS does not reliably find accurate probes are datasets that fail to isolate a single contrastive feature. We also demonstrate that Contrastive Eigenproblems are easily extended to settings with multiple features. We do so by replicating recent results which show that truth and polarity are encoded together in a shared subspace \citep{burger_truth_2024}.



\section{Understanding CCS}\label{sec:understanding_ccs}
CCS is a probing method, meaning it involves training small classifiers on the activations of a larger model in order to establish whether certain information is present.
The method yields a binary classifier, but unlike typical probes, a CCS probe is not given labels to train on.
Instead, the probe exploits its inputs consisting of contrastive pairs, which we know to have opposite feature values. 

\citet{burns_discovering_2023} focus on sentences and their negations to train the CCS probes.
For example, they used inputs that consisted of question-answer pairs like ``Is grass green? Yes/No'', or of declarative sentences such as ``Grass is green.'' and ``Grass is not green.''. 
Basically, the pair of inputs $(X^+, X^-)$ consist of language that---if uttered---would amount to an assertion ($X^+$) or denial ($X^-$) of a proposition $X$. 
The method relies only on the expectation that any latent probability distribution captured by the model's internals must sum up to one. 
Of course, we would expect such a basic consistency property for all (binary) variables, not just truth.
So in general, we have some binary \textit{feature of interest}, and $X^+$ and $X^-$ are inputs who primarily (and ideally, exclusively) differ in that feature's value. 
Typically, such inputs come in the form of minimal pairs where a word is changed, replaced or inserted in order to also change some sentence-level property.
Unless specified otherwise, we take `sentence truth' as our feature of interest.

The probes trained with CCS operate on a language model's mean-centered latent-space activations of $X^+$ and $X^-$, which we denote $\vec{x}^+$ and $\vec{x}^-$, respectively. 
In this work, we will use CCS with linear probes of the following form: $p(\vec{x}) = \sigma(\dir^\tp \vec{x})$, where $\dir$ are the probe parameters.
When using such linear probes we are assuming that there exists a direction in latent space that the language model uses to represent the feature of interest. 
By projecting activations on the \textit{feature direction} we can construct a probe that parameterizes a probability distribution for the binary feature of interest.

The objective of CCS consists of a minimization of two terms, the consistency and confidence loss:
\begin{align*} \label{eq:ccs_objective}
    \bm{\theta_{\text{ccs}}} \;=\;
    &\argmin_{\bm{\theta}} \;
    \E_{\vec{x}^+, \vec{x}^-} 
        L_\mathit{cons}^{\dir}(\vec{x}^+,\vec{x}^-)
      + L_\mathit{conf}^{\dir}(\vec{x}^+,\vec{x}^-),
\\
    \text{with \hspace{4mm}} 
    &L_\mathit{cons}^{\dir}(\vec{x}^+,\vec{x}^-) 
     \;=\;
\bigl[ 
    \sigma(\dir^\tp\vec{x}^+) - ( 1 - \sigma(\dir^\tp\vec{x}^-))
\bigr]^2
=
\bigl[ 
    \sigma(\dir^\tp\vec{x}^+) + \sigma(\dir^\tp\vec{x}^-) - 1
\bigr]^2,
\\
    \text{and \hspace{4mm}}
    &L_\mathit{conf}^{\dir}(\vec{x}^+,\vec{x}^-) 
     \;=\; 
        \min \bigl\{ \;
        \sigma(\dir^\tp\vec{x}^+), \;
        \sigma(\dir^\tp\vec{x}^-), \;
        1{-}\sigma(\dir^\tp\vec{x}^+), 
\hspace{1mm}
        1{-}\sigma(\dir^\tp\vec{x}^-)  \,
    \bigr\}^2.
\end{align*}
The consistency loss is minimized when the probabilities assigned to sentences and their negations add up to one.
The confidence loss is said to prevent the degenerate solution where $p(\vec{x}^+) = p(\vec{x}^-) = 0.5$.
We use the symmetric (unbiased) confidence loss introduced by \citet{farquhar_challenges_2023}.

\citet{burns_discovering_2023} solve this optimization problem using gradient descent, using activations for tokens from $X^+$ and $X^-$. For example, with ``Between green and blue, grass is [green/blue]'', the bracketed 
\begin{wraptable}[14]{r}{0.38\linewidth}
    \centering
    \vspace{-2mm}
    \small
    \setlength\tabcolsep{4pt}
    \caption{Mean and standard deviation of accuracy for 9 datasets, trained on activations of the (next to) last token. }
    \begin{tabular}{lrr}
        \toprule
        Dataset         &\textit{answer} (\%)&\textit{period} (\%) \\
        \midrule
        comparisons     & $100 \pm 00$    & $ 91 \pm 00$    \\
        sp\_en\_trans   & $ 99 \pm 00$    & $100 \pm 01$    \\
        cities          & $ 99 \pm 00$    & $ 99 \pm 00$    \\
        amazon          & $ 93 \pm 00$    & $ 93 \pm 00$    \\
        imdb            & $ 87 \pm 00$    & $ 87 \pm 07$    \\
        ent\_bank       & $ 79 \pm 10$    & $ 82 \pm 16$    \\
        snli            & $ 71 \pm 14$    & $ 85 \pm 06$    \\
        copa            & $ 59 \pm 06$    & $ 48 \pm 02$    \\
        rte             & $ 54 \pm 06$    & $ 54 \pm 07$    \\
        \bottomrule
    \end{tabular}
    \label{tab:base_results_9}
\end{wraptable}
tokens would be used for $\vec{x}^+$ and $\vec{x}^-$ respectively.

We begin our analysis of CCS by reporting its performance in different circumstances.
We use datasets from three sources \citep{burns_discovering_2023, marks_geometry_2024, schouten_truth-value_2025} (see \autoref{apx:datasets}). 
We include probes trained on both the last and next-to-last tokens, corresponding to a \textit{period} token and an \textit{answer} token.
We trained probes for a total of 9 datasets; 
using 30 different seeds for the probe's random initialization.

In \autoref{tab:base_results_9}, we report probe accuracy for layer 16 in Llama-2-7b \citep{touvron_llama_2023b}, which we will use throughout the paper.
We find that CCS does not reliably reach well-performing minima for all datasets. 
Specifically, there are multiple cases where the average performance is above random, but the exact performance varies between seeds.
We wonder if this could be caused by the two-termed objective, thus our next step is an investigation of what makes the two terms necessary.

\subsection{Loss-term ablations}\label{ssec:experiment1}
Given that the stated purpose of the confidence loss is to avoid the degenerate solution, it makes sense to begin by determining what other strategies could help us avoid finding that solution. 
The degenerate solution arises under the following conditions:
\[
\begin{array}{rcl}
\sigma(\bm\theta^\tp \Xp) = \sigma(\bm\theta^\tp \Xm) = 0.5 
&\enspace \implies \enspace& 
\bm\theta^\tp \Xp = \bm\theta^\tp \Xm = 0
\end{array}
\]
where $\Xp = \left[ \vec{x}_1^+, \vec{x}_2^+, \dots, \vec{x}_N^+  \right]^\tp \in \mathbb{R}^{N \times D}, $ is the data matrix for positive samples, and $\Xm$ for negative samples.
Thus, there are two ways for the degenerate case to arise:
(1) the vector learned by the probe has zero length, $|\bm\theta| = 0$; or,
(2) the direction points into the null space of the data matrix.
When training probes, it is not uncommon to work with relatively small datasets. 
Thus, the number of samples can easily be smaller than the dimensionality of the model's latent space, resulting in a rank-deficient data matrix.
%
To address both paths to the degenerate case, we can alter the CCS training process in two ways.

\paragraph{Alteration 1.}
By restricting the search space to unit vectors $\hat{\dir}$
we avoid learning the zero vector.
Note that the magnitude of the probe parameter vector is unimportant to its accuracy. 

\paragraph{Alteration 2.}
To remove the null space from the data matrix, we use singular value decomposition:
$\vec{U} \vec\Sigma \vec{V}^\tp = \mathrm{SVD}(\Xpm_{\mathit{train}})$,
where $\Xpm_\mathit{train} \in \mathbb{R}^{2N \times D}$ is a matrix containing both the  hidden states for the training data's positive samples ($\Xp_\mathit{train}$), and the negative samples ($\Xm_\mathit{train}$).
We then apply the probes to the reduced representations: $p(\vec{x}) = \sigma(\udir \vec{V}^\tp_{:r}\vec{x})$ (with $r$ being the rank of $\Xpm_\mathit{train}$).
This strategy assumes the null spaces of $\Xpm_{\mathit{train}}$, $\Xp$ and $\Xm$ are the same. 



\renewcommand{\c}[1]{\multicolumn{1}{c}{\footnotesize #1}}
\renewcommand{\L}[1]{\mathcal{L}_{\text{#1}}}

Overall, we compare the following kinds of probes:
(1) ordinary CCS;
(2) only the confidence term ($\L{conf}$); 
(3) only the consistency term ($\L{cons}$), including versions altered in one ($\L{cons}$+a1, $\L{cons}$+a2) or both ways ($\L{cons}$+a1+a2); and finally,
(4) CCS with only the alterations, no ablations (CCS+a1+a2).
For this experiment, we use the datasets for which CCS performed well in \autoref{tab:base_results_9}.

\begin{table*}[bt]
    \centering
    \small
    
    \let\mc\multicolumn
    \let\mr\multirow

    \setlength\tabcolsep{5pt}
    
    \caption{Accuracies for probes trained with ablated and/or altered objectives.}
    \begin{tabular}{p{2cm}lrrrrrrr}
    \toprule
    $\mu \pm \sigma$ (\%) && \c{CCS} & \c{$\L{conf}$} & \c{$\L{cons}$} & \c{$\L{cons}$} & \c{$\L{cons}$} & \c{$\L{cons}$} & \c{CCS}  \\
    && \c{-} & \c{-} & \c{-} & \c{+a1} & \c{+a2} & \c{+a1+a2} & \c{+a1+a2}\\
    \midrule                                                               
    \mr{3}{2cm}{\scriptsize\citet{marks_geometry_2024}}
        & comparisons      & $100{\pm}00$ &$100{\pm}01$  & $66{\pm}11$  & $62{\pm}09$ & $65{\pm}11$ & $59{\pm}07$   &$100{\pm}00$ \\
        & sp\_en\_trans    & $ 99{\pm}00$ & $96{\pm}08$  & $64{\pm}13$  & $66{\pm}12$ & $65{\pm}14$ & $74{\pm}14$   & $99{\pm}00$ \\
        & cities           & $ 99{\pm}00$ & $98{\pm}04$  & $70{\pm}14$  & $72{\pm}13$ & $77{\pm}14$ & $67{\pm}12$   & $99{\pm}00$ \\
    \cmidrule{1-9}
    \mr{2}{2cm}{\scriptsize\citet{burns_discovering_2023}}
        & amazon           & $ 93{\pm}00$ & $94{\pm}00$  & $67{\pm}09$  & $65{\pm}11$ & $72{\pm}13$ & $64{\pm}09$   & $94{\pm}00$ \\
        & imdb             & $ 87{\pm}00$ & $81{\pm}09$  & $60{\pm}06$  & $61{\pm}07$ & $63{\pm}09$ & $58{\pm}07$   & $87{\pm}01$ \\
    \bottomrule
    \end{tabular}
    \label{tab:ablation}
\end{table*}

\paragraph{Results.}
In \autoref{tab:ablation}, we give the results of the ablation.
A few things stand out:
(1) the ablation of the confidence loss-term reduces accuracy more than the ablation of the consistency loss-term, suggesting the former is more important than the latter; and
(2) the proposed alterations do not compensate for the ablation of the confidence loss-term.
These results clearly show that the role of the confidence term is not limited to preventing the degenerate solution.

\subsection{The effect of the confidence loss}
%
The confidence term encourages probabilities closer to the extremes, but what needs to be true to make that happen?
It is minimized when either the positive or negative sample of each pair are assigned a probability of zero or one. 
But, this would require that $\forall \vec{x} \,:\, \dir^\tp\vec{x} = \pm\infty$. 
Seemingly, minimizing the confidence-loss is just maximizing $||\dir^\tp \Xp||$ or $||\dir^\tp \Xm||$.

When considering unit-length vectors $\udir$, the only way to maximize $\udir^\tp \vec{x}$ is to reduce the angle between them.
This would mean that the confidence-loss is biasing $\udir$ towards directions where the data has higher variance, i.e. the first (few) principal component(s).

To test this hypothesis, we will compare the directions found by CCS to the principal components.
By ablating the two loss terms again, we can see if the confidence loss causes the direction to be more similar to the first few principal components. 
Specifically, for CCS, $\L{conf}$-only, and $\L{cons}$-only, we will measure: $
    \lambda^K(\dir) \, = \,\frac{1}{||\dir||} \left| \left|  \vec{V}_{:K} \dir \right| \right|
$. 
This measures how much of $\dir$ extends into the subspace spanned by the first $K$ principal components.

\paragraph{Results.}
\begin{wrapfigure}[14]{r}{0.5\linewidth}
    \vspace{-5mm}
    \includegraphics[
        width=0.75\linewidth, clip,
        trim=2mm 2mm 13.1cm 24.3cm,   
    ]{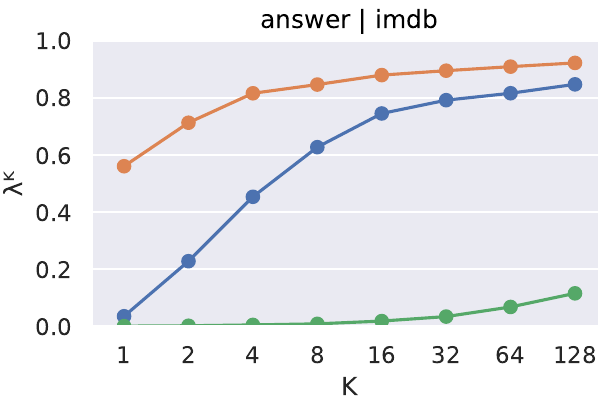}
    \includegraphics[
        scale=0.45, clip,
        trim=20.1cm 11.7cm 0 12.5cm,   
    ]{figures/conf_effect.pdf}
    \vspace{-2mm}
    \caption{Extent to which learned vectors point into the subspace spanned by the first K principal components. Shown for: CCS, only the consistency loss (--conf), and only the confdidence loss (--cons); 
    on the IMDB dataset, using activations for answer tokens, averaged over 30 random seeds.}
    \label{fig:cons_effect}
\end{wrapfigure}


%


In \autoref{fig:cons_effect}, we can see that the confidence-loss indeed causes CCS to find directions closer to the first few principal components of $\Xpm$. 
With only the consistency loss, the learned vectors have almost no magnitude in the subspace spanned by the first K principal components. 
When the confidence loss is added, its magnitude in the high-variance subspace grows.
And finally, when we train with only the confidence loss, we find vectors that on average have a cosine similarity of over 0.5 with the first principal component.
Results for other datasets and both answer and period token can be found in \autoref{apx:conf_effect}.

\paragraph{Discussion.}
But why do salient directions make for more accurate CCS probes?
We see two main reasons.
First, by virtue of how contrastive data are created, the contrastive feature is often (one of) the first principal component(s). This is especially true when using the answer token itself to probe. 
Second, what we really care about is not absolute but \underline{{relative contrast consistency}}. We have:
\begin{align*}
    \argmin_{\udir} \;   \E_{\vec{x}^+, \vec{x}^-} [ \sigma(\udir^\tp\vec{x}^+) + \sigma(\udir^\tp\vec{x}^-) - 1 ]^2 
\enspace \implies \enspace
    \argmin_{\udir} \; || \udir^\tp (\Xp + \Xm) ||,
\end{align*}
but if the variance is already low for that direction anyway, meaning $|| \udir^\tp \Xpm ||$ is already small, then the probabilities are close to $0.5$, and the high consistency is not indicative of anything meaningful.
%
The confidence loss biases CCS toward directions $\udir$ for which $||\udir^\tp \Xpm||$ is large, making it less likely that the contrast-consistency is simply due to $\udir$ pointing into a direction along which the hidden states already have low variance. However, this will always bias CCS towards high-variance directions, even when the contrastive feature is less salient. Thus, what we really want is for $|| \udir^\tp (\Xp + \Xm) ||$ to be small relative to $||\udir^\tp \Xpm||$.

\section{The Geometry of a Binary Feature}\label{sec:geometry_binary}
Before continuing to test the `relative contrast consistency' hypothesis we formulated in the previous section, we will first identify what we want from linear probes in the ideal case.


\subsection{Two Kinds of Linear}
There are (at least) two different ways to think about what is involved in learning the latent-space direction associated with a given binary feature.
On the one hand, we can think of our probe as learning to separate latent space into two regions that correspond to the possible values of the binary feature of interest.
In that case, we are learning a hyperplane's normal vector $\vec{n}$.
In the context of contrast pairs, we want to have the following property:
\begin{equation}
    \forall_i   \;:\;
    \text{sgn}(\vec{n}^\tp\vec{x}^+_i)
    \enspace = \enspace
    -\text{sgn}(\vec{n}^\tp\vec{x}^-_i).
    \label{eq:classification_property}
\end{equation}
That is, we want it to separate positive from negative samples.
This property is what we need if we want to our probe to accurately predict the feature value.
Logistic Regression is a common method to train such a probe.
And, when previous work has used classification metrics such as accuracy, they were (implicitly) treating linear probes in this way.
On the other hand, we can think of our probe as learning a direction $\vec{t}$ along which representations need to be translated for the model to treat the variable as having the opposite value. 
For contrast pairs, it is along this direction that a representation $\vec{x}^+$ would have to be translated to reach $\vec{x}^-$: 
\begin{equation}
    \forall_i    \;\;
    \exists \alpha_i                \;:\;
    \vec{x}^+_i 
    \enspace = \enspace 
    \vec{x}^-_i + \alpha_i \vec{t}.
    \label{eq:intervention_property}
\end{equation}
This is a direction we can use to intervene in a language model \citep[ActivationAddition,][]{turner_steering_2024}. 
And, when previous work \citep[e.g.][]{marks_geometry_2024} used such interventions to test if a direction models a causal variable,
they (implicitly) use this way of thinking about linear probes.

\begin{figure}[tbhp]
    \centering
    \begin{subfigure}[t]{0.49\linewidth}
        \centering
        \includegraphics[
            width=\linewidth,
            clip,
        ]{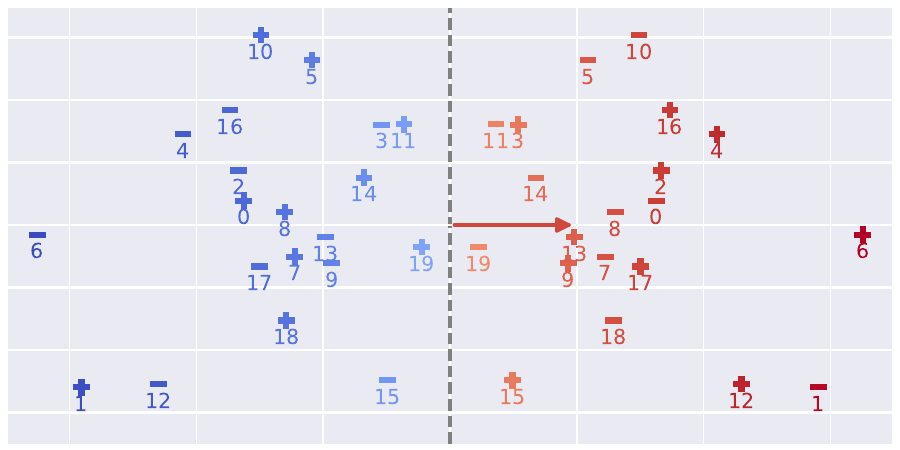}
        \caption{Ideal scenario where $\vec{x}_i^+ = \vec{x}_i^- + \alpha_i \vec{t}$, and the only other feature is represented orthogonally to $\vec{t}$.}
        \label{fig:ideal_situation}
    \end{subfigure}
    \hfill
    \begin{subfigure}[t]{0.49\linewidth}
        \centering
        \includegraphics[
            width=\linewidth,
            clip,
        ]{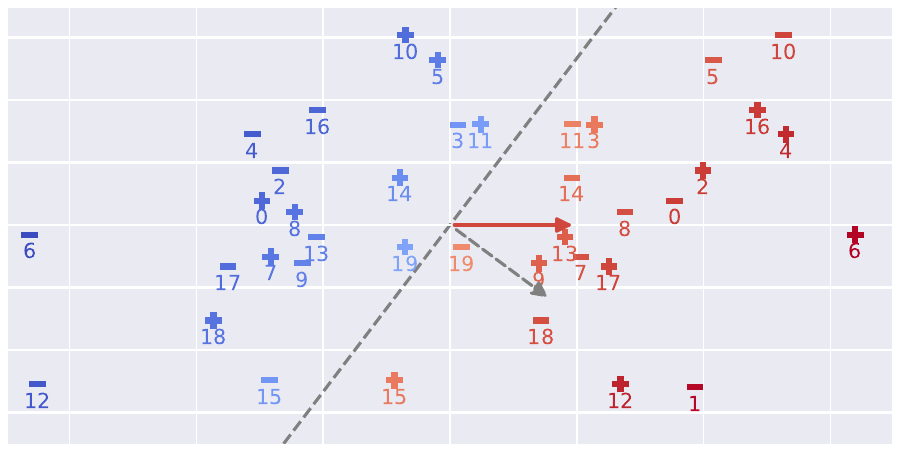}
        \caption{Scenario where $\vec{x}_i^+ = \vec{x}_i^- + \alpha_i \vec{t}$, but the only other feature is represented obliquely to $\vec{t}$. 
        The separating hyperplane's normal $\vec{n}$ is no longer the same as $\vec{t}$.}
        \label{fig:correlated_situation}
    \end{subfigure}
    \vspace{-1mm}
    \caption{Comparison of feature alignments with $\vec{t}$ in two scenarios.}
    \label{fig:both_scenarios}
\end{figure}


% ...

\subsection{Contrast Error and Displacement}
%
CCS most closely adheres to the classification-style linear probing, with its consistency loss requiring:
\begin{align*}
    \sigma(\dir^\tp \xp) = 1{-}\sigma(\dir^\tp \xm)
    \enspace \implies \enspace
    \sigma(\dir^\tp \xp) = \sigma(-\dir^\tp \xm)
    \enspace \implies \enspace
    \dir^\tp \xp = - \dir^\tp \xm.
\end{align*}
Thus, with $\dir = \vec{n}$, it requires a stronger version of the property given in \autoref{eq:classification_property}. 
Contrasting samples must not only be on opposite sides of a hyperplane but also need to be equidistant from it.
It follows that in the ideal case, we find $\dir$ in the null space of the following matrix:
\begin{align*}
    \dir^\tp ( \xm + \xp ) = 0 
    \enspace \implies \enspace
    || \dir^\tp ( \Xm + \Xp ) || = 0.
\end{align*}
We call this matrix the commonality matrix $\vec{C} = \Xm + \Xp$, since it captures the features that the positive and negative pairs have in common, i.e. the non-contrastive features that do not cancel out. 
For the intervention-style linear probing, we have:
\begin{align*}
    \xp_i + \alpha_i \vec{t} = \xm_i 
    \enspace \implies \enspace
    \Xp + \bm{\alpha} \vec{t}^\tp = \Xm 
    \enspace \implies \enspace
    \bm{\alpha} \vec{t}^\tp = \Xm - \Xp.
\end{align*}
Therefore, in order to identify $\vec{t}$ we are looking for a rank-one decomposition of $\Xm - \Xp$. 
Borrowing the terminology of \citet{fry_comparing_2023}, we will call this matrix the displacement matrix $\vec{D} = \Xm - \Xp$.

\subsection{The Ideal Case}
%
%
%
%
In \autoref{fig:both_scenarios}, we can see an idealized 2-dimensional representation how samples might be distributed in a model's latent space.
In both subfigures, the direction $\vec{t}$ is rotated onto the x-axis.
In \autoref{fig:ideal_situation}, the only other feature is uncorrelated with the binary feature of interest. 

However, 
we cannot generally assume that features are uncorrelated. 
\citet{marks_geometry_2024} point out that the feature of interest (such as sentence truth) can be correlated with other features, thereby preventing classification-style probes from finding directions like $\vec{t}$.
%
In \autoref{fig:correlated_situation}, a feature is represented in a direction {not orthogonal} to the feature of interest.
This non-orthogonal representation of the second feature, amounts to a shearing w.r.t. the situation in \autoref{fig:ideal_situation}.

For both subfigures, we can see that: 
(1) the vectors $\xm_i + \xp_i$ lie on the separating hyperplane (the dotted grey line); and,
(2) each $\xm_i - \xp_i$ lies on the x-axis. 
While in \autoref{fig:ideal_situation} the hyperplane's normal vector also lies on the x-axis, in \autoref{fig:correlated_situation} it does not.
Assuming the feature of interested is not correlated with any other features, then $\vec{t} = \vec{n}$, but in other cases the vectors can differ.

The properties we have derived for the sum and difference vectors both involve changes in variance along the direction of interest. 
When the elements of the contrast pairs are summed, the variance \textit{shrinks} (to zero for the ideal case) in the direction $\vec{n}$; and, 
when we take their difference, the variance \textit{grows} in the translation direction $\vec{t}$ (while shrinking in all other directions, to zero in the ideal case). 




\subsection{Imperfections}
In the introduction, we said ``$X^+$ and $X^-$ are inputs who primarily (and ideally, exclusively) differ in [a] feature's value''.
And it is certainly useful to pay attention to whether changes between positive and negative samples are indeed as minimal and as closely tied to the feature of interest as possible. 
However, in practice it is impossible to perfectly isolate all features this way.
It may be also be tempting to naively assume that if we did have the perfect contrastive dataset, that the properties we expect or desire from linear representations would hold exactly. 
However, besides any imperfections in the data, the model may also simply represent the data imperfectly. 
%
For both these reasons, in the next section, we will focus on finding directions with the highest increase or decrease in variance, rather than exact solutions to the equations given above.




\section{Contrastive Eigenproblems}\label{sec:eigenproblems}
In \autoref{sec:understanding_ccs}, based on our experimental results, we formulated the hypothesis that the objective of CCS amounts to finding a direction with high \textit{Relative} Contrast Consistency (RCC).
In the first part of this section, we will test this hypothesis. 
Specifically, we will approach the problem of finding a direction with high RCC as a (generalized) eigenvalue problem.
We will show that regardless of whether an such an approach is applied to consistency, or to displacement, we get the same solution in both cases.
Despite the failure to distinguish intervention- and classification-style probes, contrastive eigenproblems still have two advantages: (1) we can use the eigenvalues to get an impression of how well the dataset succeeds in isolating a single feature, and 
(2) we can straightforwardly extend the approach to probing for multiple variables.

\begin{table}[tb]
\centering
\small
\setlength\tabcolsep{4pt}
\let\mc\multicolumn
\let\mr\multirow
\let\bf\bfseries
\caption{Accuracy of DRC and RRC compared to CCS. Bold indicate results where our methods match or exceed CCS median and CRC-TPC.}
\begin{tabular}{lrrrrrrrrrrrr}
\toprule
               & \mc{6}{c}{\textit{answer}}               & \mc{6}{c}{\textit{period}} \\
               \cmidrule(lr){2-7}                       \cmidrule(lr){8-13}
               & \mc{3}{c}{CCS}  & CRC & \mr{2.2}{*}{DRC} 
                                                 & \mr{2.2}{*}{RRC} 
                                                          & \mc{3}{c}{CCS}  & CRC & \mr{2.2}{*}{DRC} 
                                                                                           &\mr{2.2}{*}{RRC}\\
Dataset        & min & med & max & -TPC&         &        & min & med & max & -TPC&        &  \\
\midrule                             
comparisons    & 100 & 100 & 100 & 100 & \bf 100 &\bf 100 & 92  & 92  & 92  &  56 & \bf 93 &\bf 94 \\
sp\_en\_trans  &  99 &  99 &  99 &  98 &     98  &    98  & 99  & 99  & 99  &  99 & \bf 99 &\bf 99 \\
cities         &  99 &  99 &  99 &  99 & \bf 99  &\bf 99  & 98  & 98  & 99  &  51 & \bf 99 &\bf 99 \\
amazon         &  94 &  94 &  94 &  94 & \bf 94  &\bf 94  & 92  & 93  & 93  &  53 & \bf 93 &\bf 93 \\
imdb           &  86 &  87 &  88 &  87 & \bf 87  &\bf 87  & 87  & 88  & 89  &  87 & \bf 89 &\bf 88 \\
ent\_bank      &  84 &  86 &  87 &  82 &     84  &\bf 86  & 48  & 93  & 94  &  86 &     89 &    90 \\
snli           &  49 &  86 &  90 &  77 &     82  &    73  & 81  & 85  & 93  &  81 & \bf 87 &\bf 87 \\
copa           &  51 &  55 &  68 &  54 &     53  &    52  & 47  & 47  & 52  &  49 &     48 &    47 \\
rte            &  46 &  50 &  61 &  49 &     50  &    50  & 45  & 57  & 68  &  58 & \bf 58 &    56 \\
\bottomrule
\end{tabular}
\label{tab:rcc_results}
\end{table}

\subsection{Problem formulation}
Based on the intuitions we developed in the last section, we now propose to solve directly for increases/decreases in variance.

\paragraph{Difference-Relative Contrast (DRC).}
Here, we express decreases/increases in variance as differences in variance between $\vec{C}$/$\vec{D}$ and $\Xpm$, giving two eigenproblems:
\begin{align*}
    \left( \vec{C}^\tp \vec{C} - \Xpm{^\tp} \Xpm \right)\vec{n}_k 
    \; = \;
    \lambda_k \vec{n}_k
    \qquad \text{and, }  \qquad  
    \left( \vec{D}^\tp \vec{D} - \Xpm{^\tp} \Xpm \right)\vec{t}_k 
    \; = \;
    \mu_k \vec{t}_k.
\end{align*}
Negative values for $\lambda_k$ correspond to directions $\vec{n}_k$ for which the variance in $\vec{C}$ is smaller than the variance in $\Xpm$. Thus, the last eigenvector of the lefthand problem is a good candidate for $\dir$.
And conversely, a positive value for $\mu_k$ indicates that $\vec{t}_k$ is a direction along which the variance in $\vec{D}$ is larger than in $\Xpm$. 
Now consider that with $\vec{A} = \Xp \pm \Xm$, both are instances of:
\begin{align*}
    \left( \vec{A}^\tp \vec{A} - \Xpm{^\tp} \Xpm \right)\vec{v}_k 
    \; = \;
    \nu_k \vec{v}_k
\\
\implies
    \left( \Xp^\tp \Xp + \Xm^\tp \Xm \pm \left( \Xp^\tp \Xm + \Xm^\tp \Xp \right) - \Xpm{^\tp} \Xpm \right)\vec{v}_k 
    \; = \;
    \nu_k \vec{v}_k
\intertext{And, because $\Xpm$ just contains the rows of $\Xp$ and $\Xm$:}
\implies
    \left( \Xp^\tp \Xm + \Xm^\tp \Xp \right)\vec{v}_k 
    \; = \;
    \pm \, \nu_k \vec{v}_k
\end{align*}
Since, both eigenproblems involve the same two cross-terms between $\Xp$ and $\Xm$; they yield the same bases in opposite order.
This means that formulating relative contrast consistency as a eigenproblem of differences in variance, forces us to assume $\vec{t} = \vec{n}$.
This approach also turns out to be very closely linked to Contrast Consistent Reflection \citep{schouten_truth-value_2025} (see \autoref{apx:connections}).

\paragraph{Ratio-Relative Contrast (RRC).}
We can also formulate generalized eigenproblems:
\begin{align*}
    \vec{C}^\tp \vec{C} \vec{n}_k
    \; = \;
    \lambda_k \Xpm{^\tp} \Xpm \vec{n}_k
    \enspace
    \implies
    \enspace
    \lambda_k 
    \; = \;
    \frac{
        \vec{n}_k^\tp \vec{C}^\tp \vec{C} \vec{n}_k
    }{
        \vec{n}^\tp_k \Xpm{^\tp} \Xpm \vec{n}_k
    },
\end{align*}
and similarly for $\vec{D}$ and $\vec{t}$.
Now the eigenvalues give ratios between the variances, rather than differences.
Both of these problems are instances of:
\begin{align*}
    \left( \Xpm^\tp \Xpm \pm \left( \Xp^\tp \Xm + \Xm^\tp \Xp \right) \right)
    \vec{w}_k
    \; &= \;
    \omega_k \Xpm{^\tp} \Xpm \vec{w}_k
\intertext{Substitute $\vec{w}_k$ for $\left(\Xpm{^\tp} \Xpm\right)^{-\frac{1}{2}}\vec{w}'_k$ and pre-multiply with $\left(\Xpm{^\tp} \Xpm\right)^{-\frac{1}{2}}$, yielding:}
    \left( \vec{I} \pm \left(\Xpm{^\tp} \Xpm\right)^{-\frac{1}{2}} \left( \Xp^\tp \Xm + \Xm^\tp \Xp \right) \left(\Xpm{^\tp} \Xpm\right)^{-\frac{1}{2}} \right)
    \vec{w}'_k
    \; &= \;
    \omega_k \vec{w}'_k
\\
    \implies 
    \left(\Xpm{^\tp} \Xpm\right)^{-\frac{1}{2}} \left( \Xp^\tp \Xm + \Xm^\tp \Xp \right) \left(\Xpm{^\tp} \Xpm\right)^{-\frac{1}{2}}
    \vec{w}'_k
    \; &= \;
    \pm (\omega_k - 1) \vec{w}'_k.
\end{align*}
Thus, this formulation too gives the same bases, and also forces $\vec{n} = \vec{t}$.

\subsection{Evaluation}
We test both approaches and compare their classification accuracy to the min/median/max accuracy of CCS (over 30 seeds).
%
The results can be seen in \autoref{tab:rcc_results}.
What is clear is that when CCS converges to the same performance reliably, both approaches match that performance almost exactly.
And, for those datasets where CCS is more sensitive to the random initialization, both approaches have accuracies somewhere between the minimum and the maximum of CCS. 
Generally both formulations perform very similarly, thus in the following experiments, we report results only for DRC.
Another method proposed by \citet{burns_discovering_2023} is CRC-TPC, which simply takes the top principal component of $\vec{D}$.
It can be seen to perform considerably worse on the period token for three datasets, likely because it finds a direction with high overall variance (see \autoref{apx:crc-tpc}).

\subsection{Interpreting Eigenvalues}
\begin{figure}
    \centering
    \begin{subfigure}[t]{0.32\linewidth}
        \includegraphics[ width=\linewidth, trim=-3mm 6mm 0 0,  ]{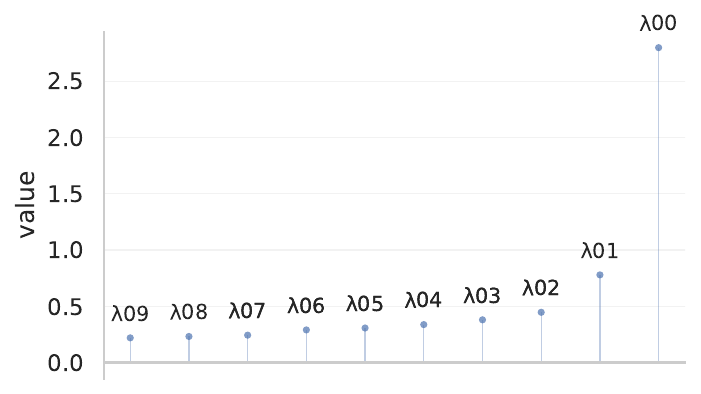}
        \caption{comparisons}
    \end{subfigure}
    \begin{subfigure}[t]{0.32\linewidth}
        \includegraphics[ width=\linewidth, trim=3mm 6mm -1mm 0,  ]{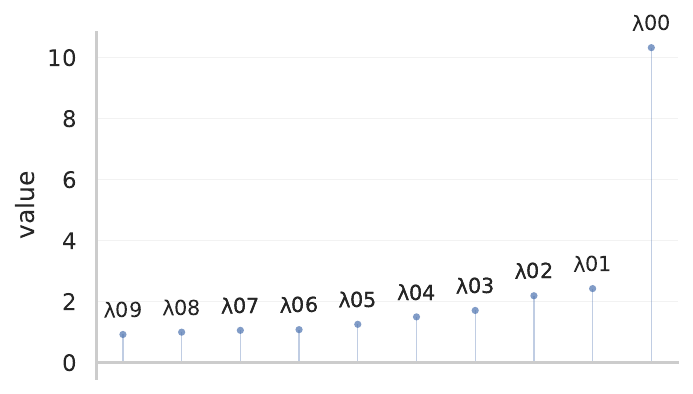}
        \caption{sp\_en\_trans}
    \end{subfigure}
    \begin{subfigure}[t]{0.32\linewidth}
        \includegraphics[ width=\linewidth, trim=0 6mm 3mm 0,  ]{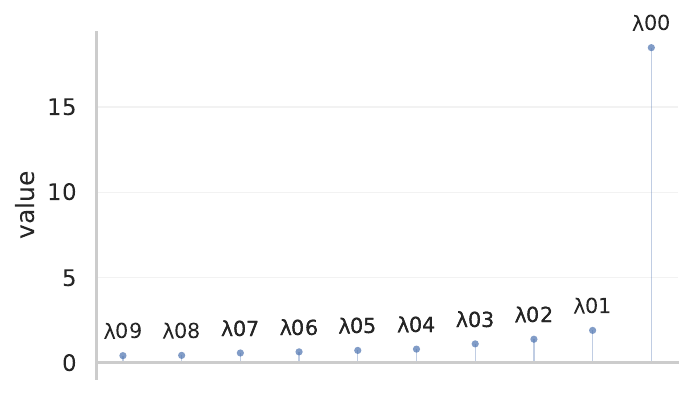}
        \caption{cities}
    \end{subfigure}
    \begin{subfigure}[t]{0.32\linewidth}
        \includegraphics[ width=\linewidth, trim=-3mm 6mm 0 0,  ]{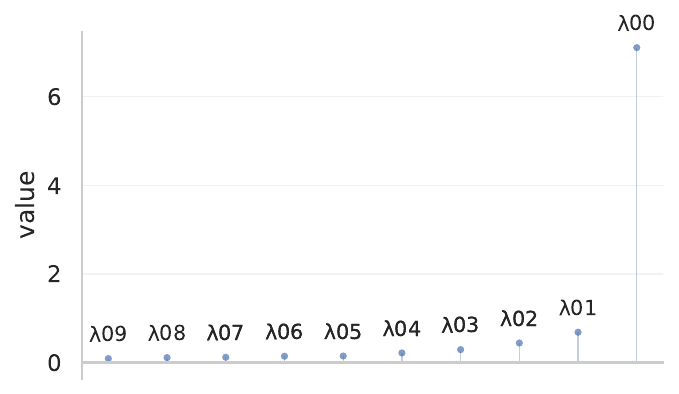}
        \caption{amazon}
    \end{subfigure}
    \begin{subfigure}[t]{0.32\linewidth}
        \includegraphics[ width=\linewidth, trim=3mm 6mm -1mm 0,  ]{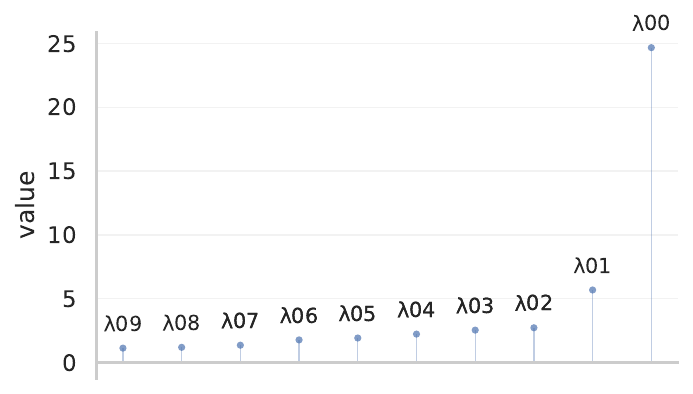}
        \caption{imdb}
    \end{subfigure}
    \begin{subfigure}[t]{0.32\linewidth}
        \includegraphics[ width=\linewidth, trim=0 6mm 3mm 0,  ]{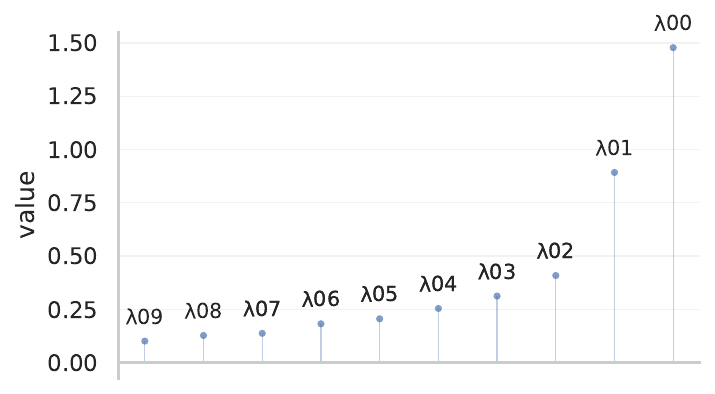}
        \caption{ent\_bank}
    \end{subfigure}
    \begin{subfigure}[t]{0.32\linewidth}
        \includegraphics[ width=\linewidth, trim=-3mm 6mm 0 0,  ]{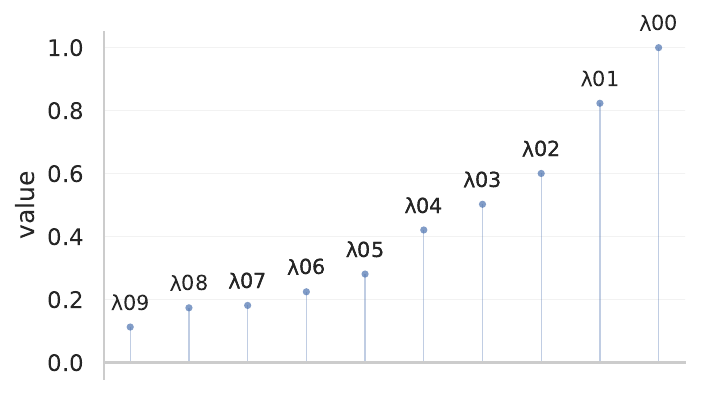}
        \caption{snli}
    \end{subfigure}
    \begin{subfigure}[t]{0.32\linewidth}
        \includegraphics[ width=\linewidth, trim=3mm 6mm -1mm 0,  ]{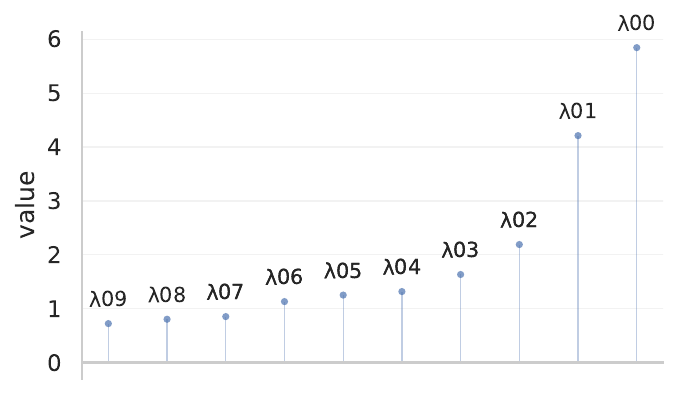}
        \caption{copa}
    \end{subfigure}
    \begin{subfigure}[t]{0.32\linewidth}
        \includegraphics[ width=\linewidth, trim=0 6mm 3mm 0,  ]{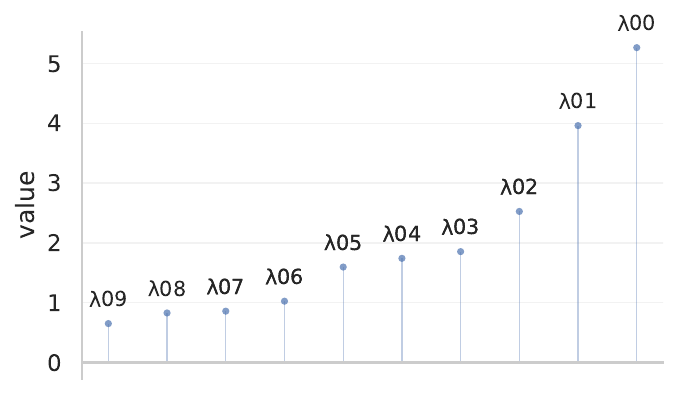}
        \caption{rte}
    \end{subfigure}
    \caption{Top DRC eigenvalues for all datasets. Based on activations taken from the answer token.}
    \label{fig:eigenvalues_interpretation_all}
\end{figure}
%
One of the benefits of approaching CCS as an eigenproblem, is that we get the whole basis of eigenvectors and their eigenvalues. 
%
One potential problem with contrast-based probing is that even if we construct the probing data ourselves, it is hard to be absolutely sure that we have truly isolated a single feature.
Not only can features be hard to differentiate from each other, but an LLM may model matters in a way that does not map onto our understanding of the problem.
Looking at the distribution of eigenvalues can be of help.
If the contrasting data and the model's representation thereof meets our expectations, then the first eigenvalue should stand out from the rest, indicating one (and only one) direction is clearly contrast-consistent.

In \autoref{fig:eigenvalues_interpretation_all}, we can see the top-10 eigenvalues for the different datasets. 
Looking at, for example, the `amazon' dataset we see precisely what we wanted, the first eigenvalue is clearly larger than the rest.
Going to `copa', we see a somewhat flatter distribution of eigenvalues. 
For this dataset, there is a second eigenvector which we will look at in more detail.
And for other datasets, like `snli', we can see an even more diffuse distribution.
Moreover, we consistently see more diffuse eigenvalues precisely in those cases where CCS's performance is less reliable (see \autoref{tab:rcc_results}).

\paragraph{Case study: COPA.}
Choice of Plausible Alternatives \citep{roemmele_choice_2011} is a commonsense causal reasoning dataset.
It consists of prompts such as: ``Consider the following example: ‘‘‘The bar closed.’’’ Choice 1: It was crowded. Choice 2: It was 3 AM. Q: Which is more likely to be the cause, choice 1 or choice 2? choice [1/2].''

\begin{table}[t]
\centering
\caption{COPA samples, showing the prompt and cause/effect options along with their activation strength $a$ on the contrastive feature encoded by DRC's top eigenvector. The choices labeled as commonsense are underlined.}
\vspace{2mm}
\let\mc\multicolumn
\let\mr\multirow
\small
\setlength\tabcolsep{2pt}
\renewcommand{\arraystretch}{1.15}
\begin{tabular}{lp{5.2cm}rlrl}
    \toprule
    & prompt & 
    \mc{1}{c}{$a$} & negative sentiment & 
    \mc{1}{c}{$a$} & positive sentiment \\ 
\midrule
    \mr{2}{*}{\rotatebox[origin=c]{90}{\underline{effect}}}
    & `The host served dinner to hist guests.'
    &  3.2 & `His guests went hungry.'
    & -2.3 & \uline{`His guests were gracious.'} \\ 
    & `A man cut in front of me in the long line.'
    &  2.3 & \uline{`I confronted him.'}
    & -2.9 & `I smiled at him.'           \\ 
\midrule
    \mr{2}{*}{\hspace{2pt}\rotatebox[origin=c]{90}{\underline{cause}}}
    & `The shirt shrunk.'
    & 1.7 & `I poured bleach on it.'
    & -2.2 & \uline{`I put it in the dryer.'} \\
    & `The smoke alarm went off.'
    &  3.1 & \uline{`I burnt my dinner.'}
    & -2.2 & `I lit a candle.'           \\ 
    \bottomrule
\end{tabular}
\label{tab:copa}
\end{table}
None of the approaches perform well on this dataset, typically doing no better than random guessing.
For the answer token, CCS does occasionally find directions that perform better at around 68\%, suggesting that the model is not at fault.
Given that the two highest eigenvalues seem to stand out among the rest, it makes sense to ask: 
(1) what does the top eigenvector represent, if not `sentence truth'?
and (2) does the second eigenvector encode `sentence truth'?

To answer the first question, we looked at a subset of COPA together with the relevant activations, i.e. the projections of the answer token hidden states onto the first eigenvector.
Looking at the most contrastive examples (where activations had the highest absolute values) quickly revealed the answer.
It appears that in COPA, `sentence truth' is not the only thing that changes between positive and negative samples, the answers often also differ in sentiment.
In \autoref{tab:copa}, we can see some examples of how high activations (left) correspond to the occurrence of comparatively `bad' situations or events in answers (negative sentiment), and low activations (right) correspond to comparatively `good' situations or events (positive sentiment). See \autoref{apx:copa_examples} for full samples.

As for the second question, the answer is likely yes. 
DRC's second eigenvector predicts `sentence truth' on COPA with 70\% accuracy, which is slightly better than the CCS's 68\% maximum.




\subsection{Multivariate Extension: Polarity and Truth}
The ability to find multiple directions can also be used deliberately.
In recent work, \citet{burger_truth_2024} show that polarity and truth occupy a shared subspace. 
To showcase the utility of our approach in a multivariate setting, we  will replicate their results.
We use the `cities' dataset, varying both the polarity (presence of negation) and the country that a city is said to lie in.
For a given city, we denote the four samples as: 
$\vec{x}^{p,c}$, $\vec{x}^{p,i}$, $\vec{x}^{n,c}$, $\vec{x}^{n,i}$, where $p$ and $n$ indicate positive and negative polarity (negation), and $c$ and $i$ indicate the correct and incorrect country.
Of these, only $\vec{x}^{p,c}$ (affirmation of correct country) and $\vec{x}^{n,i}$ (denial of incorrect country), are true statements.
Between these four points, there are six pairs to be formed.

We can use DRC on the sum of all variants to cause the variance to decrease in multiple directions:
$\vec{C} \;=\; \vec{X}^{p,c} + \vec{X}^{p,i} + \vec{X}^{n,c} + \vec{X}^{n,i}$
Equivalently, we can also concatenate the six contrast pairs, causing the variance to grow in multiple directions.
\begin{align*}
    \vec{D} \quad = \quad 
    &\begin{bmatrix}
        & \vec{X}^{p,c} - \vec{X}^{n,c} & \\
        & \vec{X}^{p,i} - \vec{X}^{n,i} & \\
        & \vec{X}^{p,c} - \vec{X}^{p,i} & \\
        & \vec{X}^{n,c} - \vec{X}^{n,i} & \\
        & \vec{X}^{p,c} - \vec{X}^{n,i} & \\
        & \vec{X}^{n,c} - \vec{X}^{p,i} & 
    \end{bmatrix}
    \quad
    \begin{matrix}
        (\text{polarity, truth}) \\
        (\text{polarity, truth}) \\
        (\text{truth}) \\
        (\text{truth}) \\
        (\text{polarity}) \\
        (\text{polarity})
    \end{matrix}
    \qquad \qquad \quad
    \Xpm  \quad = \quad
    \begin{bmatrix}
        & \vec{X}^{p,c} & \\
        & \vec{X}^{p,i} & \\
        & \vec{X}^{n,c} & \\
        & \vec{X}^{n,i} & \\
    \end{bmatrix}
    \quad
\end{align*}

\begin{figure}[h]
    \centering
    \includegraphics[
        width=0.9\linewidth, trim=6mm 4mm 3mm 3mm, 
    ]{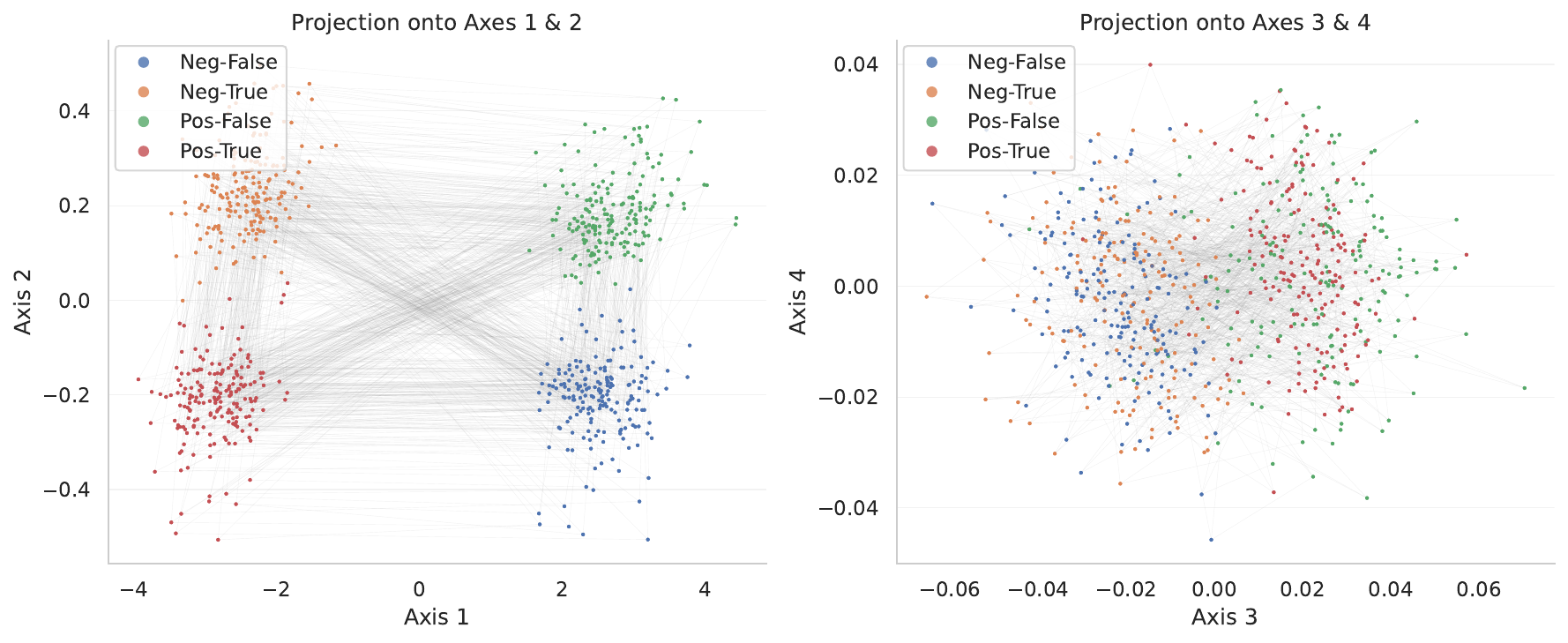}
    \vspace{-2mm}
    \caption{Projections of $\vec{x}^{p,c}$, $\vec{x}^{p,i}$, $\vec{x}^{n,c}$, $\vec{x}^{n,i}$ onto DRC's first and second eigenvectors (left) and its third and fourth eigenvectors (right). Grey lines show contrast-pairs.}
    \label{fig:multivariate_cities}
\end{figure}

\paragraph{Results.}
In \autoref{fig:multivariate_cities}, we can see activations for the portion of the cities dataset we held out for evaluation, plotted by their coordinates along the first/second, and third/fourth eigenvectors found by DRC.
We see clear separation between true and false statements by the first eigenvector.
The second eigenvector separates statements by the truth of the base (unnegated) proposition.
Finally, the third eigenvector separates statements by their polarity.
Our results clearly show that there are three orthogonal directions (the first three eigenvectors) in Llama-2 that together encode truth and polarity.

\section{Related Work}
CCS has seen a number of other analyses since its publication.
\citet{fry_comparing_2023} introduce the midpoint-displacement loss function.
This function uses the same sum and difference vectors that make up our $\vec{C}$ and $\vec{D}$ matrices. 
They argue that CCS is optimizing for a trade-off between the angle to the sum vectors and the angle to the difference vectors.
We argue that both should be relativized, removing the component that biases CCS to the directions of higher variance.
\citet{farquhar_challenges_2023} give proofs and demonstrate empirically that CCS can find features other than truth. 
While this can get in the way of the original goal of `eliciting latent knowledge', it can also be seen as an advantage that makes CCS more widely applicable.
Our approach gives an orthonormal basis with eigenvalues that indicate to what extent each direction is contrast consistent. 
Thus, if there are multiple binary features present in our contrast pairs, this can be diagnosed and, if necessary, addressed.
\citet{levinstein_still_2024} also identify problems around isolating truth, and report CCS failing to learn the sentence truth feature under various experimental settings, possibly because it was learning the another feature instead. 
\citet{belrose_vinc-s_2024} analyze and extend CRC-TPC, a closely related method. 
They show how the objective of CRC-TPC can be decomposed into separate interpretable terms, and then propose a number of additions, including a paraphrase invariance term and a supervised term. 
Similar to this work, their approach is also reducible to an eigendecomposition.
While the solutions explored are similar, \citeauthor{belrose_vinc-s_2024} focus on classification performance, while we explore how formulating contrastive eigenproblems can help diagnose problems in the data, and enable extensions to multivariate settings.
Finally, \citet{laurito_cluster-norm_2024} point to a problem where directions may be found that encode functions of features already represented along their own direction (e.g. the polarity-sensitive truth direction). 
To address it, they introduce a procedure where activations are clustered and then normalized (on both mean and variance) within each cluster. 
They show that this procedure improves the accuracy of CCS and CRC-TPC in various settings.
While \citeauthor{laurito_cluster-norm_2024} try to eliminate less-relevant contrastive directions, we aim to find all of them.




\section{Conclusion}
We have explored: (1) how CCS functions; (2) what linear probes should learn in the ideal case; and (3) how CCS might be formulated as an eigenproblem and the advantages of doing so.
We have argued that the confidence loss is an imperfect way of ensuring CCS probes find directions with high \textit{relative} contrast consistency. 
We identified two ways of thinking about linear probes (classification-style and intervention-style) and how contrastive data can help to find them.
In trying to solve for such probes by formulating eigenproblems, we have not succeeded in identifying distinct methods to solve for one of the two types of linear probes.
However, what the eigenproblem approach does provide is: 
(1) interpretable eigenvalues that indicate how well our contrastive data isolates a single feature, and 
(2) a natural extension to the multivariate setting.
Looking at the eigenvalues, we have seen that large variance in CCS's performance can be explained by datasets failing to isolate one and only one contrastive feature.
Using multivariate contrastive eigenproblems, we have replicated recent results showing how truth and polarity are encoded in the latent spaces of language models.
We believe these results show that Contrastive Eigenproblems provide a useful tool. It either yields an accurate probe, or the means to explain why such a probe is difficult to find for a particular dataset. 
Future work should look for contrastive probing techniques that can find separate directions which are optimal for either classification or intervention.

\begin{ack}
    This research was supported by Huawei Finland through the DreamsLab project. All content represented the opinions of the authors, which were not necessarily shared or endorsed by their respective employers and/ or sponsors.
\end{ack}



\bibliography{references}
\bibliographystyle{abbrvnat}

\newpage
\appendix
\section{Maximum variance effect of confidence-loss}
\label{apx:conf_effect}
\begin{figure*}[h!]
    \centering
    \includegraphics[width=\linewidth]{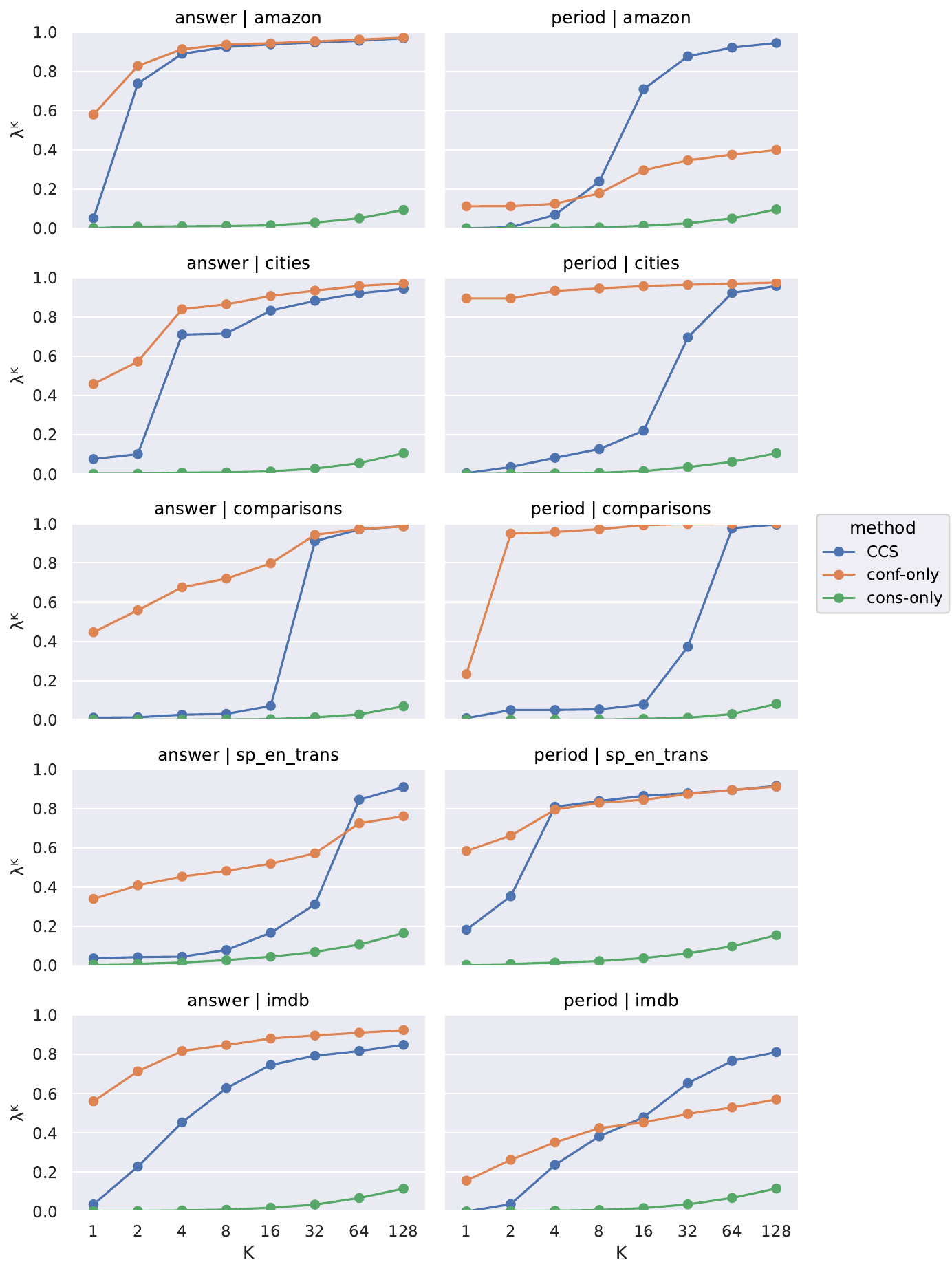}
    \caption{Effect of loss terms on probe parameter vector's similarity to top principal components.}
    \label{fig:conf_effect_full_normalized}
\end{figure*}
\FloatBarrier
\newpage
\section{Performance of CRC on top 5 contrastive principal components} \label{apx:crc-tpc}
Here we show the performance of the principal components of $X^- - X^+$.
When we use the top principal component this is equivalent to CRC-TPC \citep{burns_discovering_2023}
\begin{table*}[ht]
    \let\b\bfseries
    \let\c\mathbf
    \def\s{\itshape}
    \centering
    \caption{Accuracy of classifying with principal components of $\Xp - \Xm$ for 9 datasets. }
    \begin{tabular}{llrrrrr}
\toprule
Dataset       & Token  & PC1    & PC2   & PC3   & PC4   & PC5  \\ 
\midrule
comparisons   & answer &\b 1.00 &   .47 &   .53 &   .45 &   .57 \\
sp\_en\_trans & answer &\s  .98 &   .45 &   .61 &   .46 &   .67 \\
cities        & answer &\b  .99 &   .63 &   .55 &   .41 &   .55 \\
amazon        & answer &\b  .94 &   .47 &   .51 &   .50 &   .54 \\
imdb          & answer &\b  .87 &   .58 &   .51 &   .51 &   .52 \\
ent\_bank     & answer &\s  .82 &   .54 &   .48 &   .58 &   .51 \\
snli          & answer &\b  .77 &   .59 &\b .77 &   .60 &   .56 \\
copa          & answer &    .54 &\b .71 &   .52 &   .60 &   .50 \\
rte           & answer &    .49 &\b .68 &   .60 &   .53 &   .58 \\ 
\midrule
comparisons   & period &    .56 &   .52 &   .56 &\b .94 &   .55 \\
sp\_en\_trans & period &\b  .99 &   .51 &   .56 &   .58 &   .61 \\
cities        & period &    .51 &\b .97 &   .58 &   .46 &   .59 \\
amazon        & period &    .53 &\b .92 &   .49 &   .54 &   .50 \\
imdb          & period &\b  .87 &   .51 &   .52 &   .49 &   .53 \\
ent\_bank     & period &\b  .86 &   .51 &   .61 &   .59 &   .51 \\
snli          & period &\b  .81 &   .67 &   .51 &   .58 &   .50 \\
copa          & period &    .49 &   .46 &   .53 &   .48 &\b .65 \\
rte           & period &    .58 &\b .66 &   .51 &   .59 &   .51 \\
\bottomrule
    \end{tabular}
    \label{tab:ccs_vs_principal_components}
\end{table*}

\def\cov{\text{Cov}}

\section{Connections between DRC, CCR, and CRC-TPC}\label{apx:connections}
\citet{belrose_vinc-s_2024} helpfully point out that CRC-TPC can be broken down as follows.
They remind us that: $\text{Var}(A - B) = \text{Var}(A) + \text{Var}(B) - 2\cov(A, B)$.
Meaning the top principal component $\vec{w}^*$ of $\vec{D}$ can be written as:
\begin{align*}
\vec{w}^*
&= 
\argmax_{||\vec{w}||_2=1} \; 
\vec{w}^\tp \cov(\vec{D}) \vec{w}
\\&= 
\argmax_{||\vec{w}||_2=1} \; 
\vec{w}^\tp \cov(\Xp - \Xm) \vec{w}
\\
&=
\argmax_{||\vec{w}||_2=1} \; 
\vec{w}^\tp \left( \cov(\Xp) + \cov(\Xm) - \cov(\Xp, \Xm) - \cov(\Xm, \Xp) \right) \vec{w},
\intertext{where $\cov(\cdot, \cdot)$ denotes the cross-variance. Using notation from \autoref{sec:understanding_ccs}, we have:}
&=
\argmax_{||\vec{w}||_2=1} \; 
\vec{w}^\tp \left( 2 \cov(\Xpm) - \cov(\Xp, \Xm) - \cov(\Xm, \Xp) \right) \vec{w}
\\
&=
\argmax_{||\vec{w}||_2=1} \; 
\vec{w}^\tp \left( \Xpm{^\tp}\Xpm - \Xp{^\tp}\Xm - \Xm{^\tp}\Xp \right) \vec{w}.
\end{align*}

\citet{schouten_truth-value_2025} introduce Contrast Consistent Reflection.
They note that the objective of CCS requires that a pair of contrasting activations lie on opposite sides of, and equidistant from, a hyperplane.
They propose that it may be beneficial to train probes that require points to also be each other's exact reflection through the hyperplane.
Their proposed objective is:
\begin{align*}
\vec{r}^*
&=
\argmin_{||\vec{r}||_2=1} \; 
\mathbb{E}_{\xp, \xm} 
|| \xp - \left( \vec{I} - 2 \vec{r}\vec{r}^\tp \right) \xm ||_2,
\intertext{which is equivalent to:}
&=
\argmin_{||\vec{r}||_2=1} \; 
|| \Xp{^\tp} - \left( \vec{I} - 2 \vec{r}\vec{r}^\tp \right) \Xm{^\tp} ||_{2,1}.
\intertext{With a Frobenius norm, we change the objective slightly, but allow for a closed-form solution. }
\vec{r}^*
&=
\argmin_{||\vec{r}||_2=1} \; 
|| \Xp{^\tp} - \left( \vec{I} - 2 \vec{r}\vec{r}^\tp \right) \Xm{^\tp} ||_F^2
\\
&=
\argmin_{||\vec{r}||_2=1} \; 
|| \Xp{^\tp} - \Xm{^\tp} + 2 \vec{r} \left(\vec{r}^\tp \Xm{^\tp} \right) ||_F^2
\intertext{With $||\vec{A}+\vec{B}||_F = ||\vec{A}||_F + ||\vec{B}||_F + 2\,\text{tr}(\vec{A}^\tp \vec{B})$, we have:}
&=
\argmin_{||\vec{r}||_2=1} \; 
|| \Xp{^\tp} - \Xm{^\tp} ||_F^2 
+ 4 ||\vec{r} \left(\vec{r}^\tp \Xm{^\tp} \right)||_F^2 
+ 4 \, \text{tr}((\Xp{^\tp} - \Xm{^\tp})^\tp \vec{r}(\vec{r}^\tp \Xm{^\tp}))
\\
&=
\argmin_{||\vec{r}||_2=1} \; 
4 \vec{r}^\tp \Xm{^\tp} \Xm \vec{r} 
+ 4 \, \vec{r}^\tp \Xm{^\tp}(\Xp{^\tp} - \Xm{^\tp})^\tp \vec{r}
\\
&=
\argmin_{||\vec{r}||_2=1} \; 
\vec{r}^\tp \Xm{^\tp} \Xp \vec{r} 
\intertext{And, because the quadratic form only depends on the symmetric part:}
&=
\argmin_{||\vec{r}||_2=1} \; 
\vec{r}^\tp \left( \Xm{^\tp}\Xp + \Xp{^\tp}\Xm \right) \vec{r} 
\\
&=
\argmax_{||\vec{r}||_2=1} \; 
\vec{r}^\tp \left(- \Xm{^\tp}\Xp - \Xp{^\tp}\Xm \right) \vec{r} 
\end{align*}
Which is identical to both: (1) the terms that the cross-covariance terms contributed to the derivation for CRC-TPC; and, (2) the objective for the first (or last) eigenvector of DRC as shown in \autoref{sec:eigenproblems}.
\newpage
\section{Full COPA examples}\label{apx:copa_examples}
\begin{table}[h]
\small
\centering
\caption{Strongly and weakly activating samples in COPA for DRC's first eigenvector. The answer choice corresponding to the high value is highlighted in bold. }
\begin{tabular}{rrp{11cm}}
\toprule
\multicolumn{2}{c}{Act. strengths} 
                        &    Prompt \\
\midrule
 3.06  &  -3.49    &    Consider the following example: ‘‘‘I finished a page of the book.’’’ Choice 1: \textbf{I ripped out the next page.} Choice 2: I turned to the next page. Q: Which one is more likely to be the effect, choice 1 or choice 2? choice [1/2].\\
 3.23  &  -2.25    &    Consider the following example: ‘‘‘The host served dinner to his guests.’’’ Choice 1: His guests were gracious. Choice 2: \textbf{His guests went hungry.} Q: Which one is more likely to be the effect, choice 1 or choice 2? choice [1/2].\\
 3.28  &  -1.59    &    Consider the following example: ‘‘‘The man contemplated the painting.’’’ Choice 1: He felt in awe. Choice 2: \textbf{He collapsed.} Q: Which one is more likely to be the effect, choice 1 or choice 2?  choice [1/2].\\
 3.32  &  -2.48    &    Consider the following example: ‘‘‘The woman filed a restraining order against the man.’’’ Choice 1: The man called her. Choice 2: \textbf{The man stalked her.} Q: Which one is more likely to be the cause, choice 1 or choice 2?  choice [1/2].\\
 3.06  &  -2.18    &    Consider the following example: ‘‘‘The smoke alarm went off.’’’ Choice 1: I lit a candle. Choice 2: \textbf{I burnt my dinner.} Q: Which one is more likely to be the cause, choice 1 or choice 2?  choice [1/2].\\
 3.91  &  -2.94    &    Consider the following example: ‘‘‘The scientist conducted an experiment.’’’ Choice 1: She validated her theory. Choice 2: \textbf{She fabricated her data.} Q: Which one is more likely to be the effect, choice 1 or choice 2? choice [1/2].\\
 2.58  &  -3.41    &    Consider the following example: ‘‘‘The girl desired her parent's approval.’’’ Choice 1: \textbf{She ran away from home.} Choice 2: She obeyed her parent's rules. Q: Which one is more likely to be the effect, choice 1 or choice 2? choice [1/2].\\
 2.66  &  -3.35    &    Consider the following example: ‘‘‘The detective flashed his badge to the police officer.’’’ Choice 1: \textbf{The police officer confiscated the detective's badge.} Choice 2: The police officer let the detective enter the crime scene. Q: Which one is more likely to be the effect, choice 1 or choice 2? choice [1/2].\\
 2.29  &  -2.94    &    Consider the following example: ‘‘‘A man cut in front of me in the long line.’’’ Choice 1: \textbf{I confronted him.} Choice 2: I smiled at him. Q: Which one is more likely to be the effect, choice 1 or choice 2? choice [1/2].\\
%
1.69 & -2.16 &  Consider the following example: ‘‘‘The shirt shrunk.’’’ Choice 1: \textbf{I poured bleach on it.} Choice 2: I put it in the dryer. Q: Which one is more likely to be the cause, choice 1 or choice 2? choice [1/2].\\
 \midrule
 0.22  &   0.13	    &    Consider the following example: ‘‘‘A burglar broke into the house.’’’ Choice 1: The homeowners were asleep. Choice 2: The security alarm went off. Q: Which one is more likely to be the effect, choice 1 or choice 2? choice [1/2].\\
 0.04  &   0.32	    &    Consider the following example: ‘‘‘The baby was wailing in his crib.’’’ Choice 1: The mother picked up the baby. Choice 2: The baby crawled to the mother. Q: Which one is more likely to be the effect, choice 1 or choice 2?  choice [1/2].\\
 0.04  &   0.33	    &    Consider the following example: ‘‘‘I pushed the gas pedal.’’’ Choice 1: The car accelerated. Choice 2: The car door opened. Q: Which one is more likely to be the effect, choice 1 or choice 2? choice [1/2].\\
-0.10  &  -0.32    &    Consider the following example: ‘‘‘The investigators deemed the man's death a suicide.’’’ Choice 1: He left a note. Choice 2: He had children. Q: Which one is more likely to be the cause, choice 1 or choice 2? choice [1/2].\\
-0.16  &  -0.35    &    Consider the following example: ‘‘‘The girl performed in a dance recital.’’’ Choice 1: Her parents showed her how to dance. Choice 2: Her parents came to watch the recital. Q: Which one is more likely to be the effect, choice 1 or choice 2?  choice [1/2].\\
 0.34  &  -0.58    &    Consider the following example: ‘‘‘The man was bitten by mosquitoes.’’’ Choice 1: He went camping in the woods. Choice 2: He fell asleep on his couch. Q: Which one is more likely to be the cause, choice 1 or choice 2? choice [1/2].\\
\bottomrule 
\end{tabular}
\end{table}
\newpage
\section{Datasets}\label{apx:datasets}

\subsection{Prompts used for original CCS datasets}

\lstdefinestyle{samples}{
  breaklines, 
  breakatwhitespace=true,
  basicstyle=\fontfamily{cmr}\fontseries{sc}\selectfont\footnotesize,
  columns=fullflexible
}
\lstset{
    numbers=left,
    escapechar=@,
    style=samples,
    literate=
        {{[}in{]}}{{\CodeSymbol{in}}}4
    ,
}

\subsubsection{amazon}
\begin{lstlisting}
{
    answer_choices: "negative ||| positive",
    jinja: "Consider the following example: ''' {{content}} ''' Between {{answer_choices[0]}} and {{answer_choices[1]}}, the sentiment of this example is ||| {{answer_choices[label]}}"
}
\end{lstlisting}

\subsubsection{imdb}
\begin{lstlisting}
{
    answer_choices: "negative ||| positive",
    jinja: "The following movie review expresses what sentiment? {{text}} ||| {{answer_choices[label]}}"
}
\end{lstlisting}

\subsubsection{copa}
\begin{lstlisting}
{
    answer_choices: "choice 1 ||| choice 2",
    jinja: "Consider the following example: '''{{premise}}''' Choice 1: {{choice1}} Choice 2: {{choice2}} Q: Which one is more likely to be the {{question}}, choice 1 or choice 2? ||| {{answer_choices[label]}}"
}
\end{lstlisting}

\subsubsection{rte}
\begin{lstlisting}
{
    answer_choices: "incorrect ||| correct",
    jinja: 'Assuming that the following is true: "{{text1}}"\nConcluding that: "{{text2}}" is ||| {{answer_choices[label]}}'
},
\end{lstlisting}


\newpage

\end{document}